\documentclass[conference]{IEEEtran}
\IEEEoverridecommandlockouts
\usepackage{cite}
\usepackage{amsmath,amssymb,amsfonts}
\usepackage{url}

\usepackage{graphicx}
\usepackage{textcomp}
\usepackage{xcolor}

\usepackage{algorithmicx}
\usepackage{algpseudocode}
\usepackage{algorithm}

\def\BibTeX{{\rm B\kern-.05em{\sc i\kern-.025em b}\kern-.08em
    T\kern-.1667em\lower.7ex\hbox{E}\kern-.125emX}}

\begin{document}

\title{Using LLMs to discover emerging {\it coded} antisemitic hate-speech in extremist social media\\
\thanks{Research supported by American University's Signature Research Initiative program. We thank Jacob Levine for his inspiration in the initial steps of this project.}
}

 \author{\IEEEauthorblockN{Dhanush Kikkisetti, Raza Ul Mustafa, Wendy Melillo, Roberto Corizzo, Zois Boukouvalas, Jeff Gill, Nathalie Japkowicz}
  \IEEEauthorblockA{\textit{American University, 4400 Massachusetts Ave NW, Washington, DC 20016, USA} \\
 \{vk4372a,rmustafa,melillo,rcorizzo,boukouva,jgill,japkowic\}@american.edu}
 }
\maketitle

\begin{abstract}
Online hate speech proliferation has created a difficult problem for social media platforms. A particular challenge relates to the use of coded language by groups interested in both creating a sense of belonging for its users and evading detection. Coded language evolves quickly and its use varies over time. This paper proposes a methodology for detecting emerging coded hate-laden terminology.
The methodology is tested in the context of online antisemitic discourse. The approach considers posts scraped from social media platforms, often used by extremist users. The posts are scraped using seed expressions related to previously known discourse of hatred towards Jews.
%
%
The method begins by identifying the expressions most representative of each post and calculating their frequency in the whole corpus. It filters out grammatically incoherent expressions as well as previously encountered ones so as to focus on emergent well-formed terminology. This is followed by an assessment of semantic similarity 
to known antisemitic terminology using a fine-tuned large language model, and subsequent filtering out of the expressions that are too distant from known expressions of hatred. Emergent antisemitic expressions containing terms clearly relating to Jewish topics are then removed to return only coded expressions of hatred. 
\end{abstract}


\section{Introduction}
\begin{figure}
    \centering
    \includegraphics[width=0.75\linewidth]{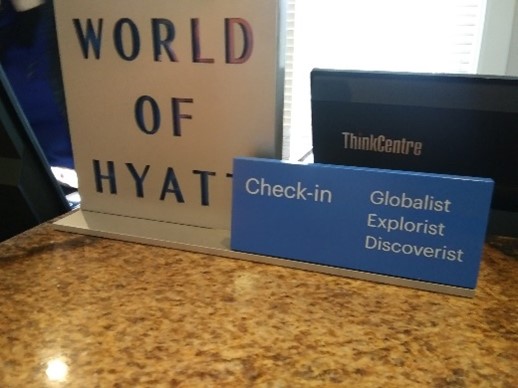}
    \caption{Non antisemitic use of the term "globalist"}
    \label{fig:globalists}
\end{figure}

Online hate speech detection\footnote{Warning: Some of the 
paper's content 
may be disturbing to the reader.} is a complex problem for social media platforms. A particular challenge, not much discussed in the literature, relates to the use of coded language. 
The following post illustrates the issue in the context of antisemitic hate speech:  
\begin{quote}
“Nope Globalist want us intertwined and run by the elites, Globalist don't lay tariffs on their friends you stupid fu****”. [posted on Dec. 31, 2022, on the Disqus platform 
]
\end{quote}
According to the American Jewish Committee (AJC) Translate Hate Glossary\footnote{\url{https://www.ajc.org/translatehate/globalist}}, a {\it globalist}, in its unbiased definition, is “a person who advocates the interpretation or planning of economic and foreign policy in relation to events and developments throughout the world”. According to this definition, the term is rather flattering. Indeed, that is the way it is intended in the Hyatt hotel’s welcoming message to its club members seen in Figure~\ref{fig:globalists}\footnote{Photo by one of the authors on 11/3/23 at a Hyatt Texas property.}. In that commercial context a globalist refers to someone ``who gets it!" and should feel good about it! The term does not, in any way, refer to Jews.
Yet,  the AJC Translate Hate Glossary argues that the term has an antisemitic connotation when it “is used to promote the antisemitic conspiracy that Jewish people do not have allegiance to their countries of origin, like the United States, but to some worldwide order—like a global economy or international political system—that will enhance their control over the world’s banks, governments, and media”. 
In the above post, it is clear that 
the antisemitic connotation is implied. From this post, we surmise that
\begin{enumerate}
    \item The globalists (a.k.a., the Jews) are distinct from “us”, presumably, the good American citizens;
    \item They control “our” fate to be run by the elites (a subset of these Jews)\footnote{``Elite" appears in the AJC Glossary in the context of ``Cosmpolitan Elite": ````Cosmopolitan” and ``elite” are terms that have separately incited antisemites across the political spectrum. Based on stereotypes of Jewish wealth and insularity, Jews have been accused of being part of an elite class for centuries."};
    \item They help each other by not imposing the same tariffs on each other as those they impose on ``us".  
\end{enumerate}
The above post, thus has a double meaning. To a recipient who is unaware of its antisemitic connotation, some category of people, the {\it globalists}, do not seem to behave very nicely. Yet, to an informed audience, it is a very pointed post that reiterates old Nazi and Soviet anitisemitic propaganda\footnote{c.f. ``Globalist" and ``Cosmopolitan Elite" in the AJC Glossary.} and propagates it further. Furthermore, on social media platforms, it does so without setting off any serious alerts since, except for the “stupid fu****” mention, which could raise a flag, no offensive terms are used.

Though the usefulness and importance of catching such subtle posts and their impact on society beyond the small extremist groups they are primarily intended for are important subjects that we debate elsewhere, this paper is concerned with the automatic discovery of ``coded" terms similar to {\it globalists} and {\it cosmopolitan elite} which carry both a ``regular" and an antisemitic connotation depending on the context in which they are used. Such an automated process is necessary due to the fact that coded terminology evolves rapidly online and fixed glossaries such as the AJC glossary become quickly outdated. In addition, due to the large volume of posts appearing on social media, human monitoring cannot be performed without the assistance of automated tools pointing them in the right direction. The purpose of our approach is just that: to create an automated monitoring tool to assist human monitors by suggesting emerging, potentially coded, antisemitic terminology, along with the posts that use that terminology. 

Though the topic of hate speech is, unfortunately, quite vast, this study focuses on antisemitism. The choice of a particular category of hate speech comes from our belief that we can perform a more thorough analysis of the problem by remaining focused. Antisemitism was selected because of the reported increase in antisemitic incidents in the months preceding the beginning of this study, in 2022. 
The lessons learned from this particular study will apply 
to other categories of hatred including hatred against Black, Muslim, Asian, and LGPTQ+ populations amongst others.  

The main contribution of this paper is a methodology for the novel problem of extracting {\it emerging coded hate-laden terminology} (antisemitism, in this paper) from extremist social posts, along with a practical pipeline to demonstrate its effectiveness. 
The methodology is based on the  hypothesis that
{\bf coded antisemitic terminology begets coded antisemitic terminology.}       
In other words, those who use coded terminology to remain under the radar of social media monitors will, when not able to express new ideas with existing coded terms, derive or invent new ones.
Based on this hypothesis, we harvest terminology used in similar contexts as known coded antisemitic terminology and propose it as potential emerging antisemitic coded terminology to human monitors, along with the context in which that terminology occurs. 
We propose four different versions of our pipeline and validate them using a quantitative approach. The most advanced version is also evaluated qualitatively. We conclude with a discussion of our approach's practical utility. 

The remainder of the paper is structured as follows: Section \ref{tab:background} presents background and related work. In Section \ref{tab:fw}, we discuss data preparation matters. Next, the methodology and pipeline for extracting coded terminology is introduced in detail in Section \ref{approach}. 
This is followed by a presentation and discussion of the results in Section \ref{tab:results_discussion}. Finally, Section \ref{tab:conclusion} concludes the paper and discusses future work.
\vspace{-0.2cm}
%
%
\section{Background and related work} \label{tab:background}

With the advent of the internet and social media, technology has increased the speed at which language evolves. Propaganda in the form of hate speech now travels the world at such a fast pace that it is beyond human capacity to keep up with. Harmful words take on new meanings in both direct and coded ways, inciting hatred in the minds of those only too willing to believe them as they reinforce and justify preexisting prejudices. 

\subsection{Machine Learning Methods for Hate Speech Detection}
In recent times, there has been a notable rise in hate crimes across the United States.\footnote{\url{https://bjs.ojp.gov/library/publications/hate-crime-recorded-law-enforcement-2010-2019}} While establishing a clear relationship between hate crimes and online content is not straightforward, a 
report by the US Department of Justice points to the simultaneous purchases of Facebook ads containing dividing content and hate crime. These two reports\footnote{(1) \url{https://bit.ly/2xeeF5h}; (2) \url{https://www.ojp.gov/pdffiles1/nij/grants/304532.pdf}}  thus suggest that hate speech should not be considered harmless, and coming up with methods to curb it is an important goal.  

Previous work aims to detect hate speech from social media using various Machine Learning (ML) methods as documented by a number of surveys written in the last six years \cite{Schmidt2017ASO, Fortuna2018ASO, Poletto2020ResourcesAB, Jahan2021ASR}. One of the most recent surveys shows that while up to 2016, fewer than 10 papers were published on the topic each year, since then, there has been a huge increase in interest in the topic with over 150 papers published in 2020, the last year for which their survey had complete information \cite{Jahan2021ASR}. Hate speech detection has been attempted using a wide variety of techniques and applied to many different problems.  Founta et al.\cite{founta2019unified}, for example, used Recurrent Neural Networks (RNN) to classify racism and sexism. Serrà et al. \cite{serra2017class} showed that character level based Long Short-Term Memory networks (LSTMs) for abusive language detection could be useful. Similarly, Convolutional Neural Networks (CNNs) have also been shown to be successful in hate speech detection and classification \cite{gamback2017using}. More recently, large language models have been used for these tasks like in the work of \cite{Wiedemann2020UHHLTL} who propose different fine-tuned and non-fined-tuned variations of pre-trained models such as BERT, RoBERTa, ALBERT, etc. on offensive language detection. Most of these studies, however, consider hate speech as a whole and, typically, do not distinguish the community towards which it is directed. We feel that this generalized approach is too broad and decided, instead, to use a divide-and-conquer approach by focusing on particular communities separately. Our first attempt focused on the Jewish community and the problem of antisemitic speech in social media. 
\subsection{Antisemitism in Social Media and its Detection}

Antisemitism specifically targets Jewish individuals or the Jewish community \cite{schwarz2017inside}. 
In \cite{zannettou2020quantitative}, authors use the outcomes of two surveys from EU and ADL to assess how
the level of antisemitism relates to the perception of antisemitism by the Jewish community in eight different EU countries.
A recent survey finds that 20\% of American Jewish adults have experienced an act of antisemitism, such as an attack either online or on social media.\footnote{\url{https://bit.ly/41FV6ei}} In another study, the authors address the challenges of quantifying and measuring online antisemitism. 
It raises the question of whether the number of antisemitic messages is increasing proportionally to other content or if the share of antisemitic content is rising. Additionally, the paper aims to determine the extent of online Jew-hatred beyond well-known websites, forums, and closed social media groups \cite{jikeli2019annotating}.\footnote{These studies preceded 10/7/23 when the situation worsened drastically.}

A few studies have attempted to combat online antisemitism in a way similar to the way in which generalized hate speech has been countered in the works discussed in the previous section. In \cite{Chandra2021SubvertingTJ}, for example, the authors prepared a data set that includes both social posts and associated images, when available. They labeled the entries as antisemitic or not, and if antisemitic, indicated the kind of antisemitism: political, economic, religious or racial. They used a bimodal deep learning approach for classifying the data into these categories. \cite{Cloutier2023} considers a subset of the text-only part of this dataset
in an attempt to classify antisemitic posts using a less computationally-intensive approach. Focusing on the class imbalance problem in the data while taking advantage of OpenAI's GPT technology, they compared GPT-based resampling techniques against other traditional kinds. 
Very recently, \cite{Jikeli2023AntisemiticMA} proposed a new data set for antisemitism detection in social media posts that uses a strict annotating process. The data set is so recent, however, that it has not yet been used for classification or the results obtained on such efforts have not yet been published.
There are other projects that consider the detection of online antisemitism using AI approaches as well. In particular, the project entitled ``Decoding Antisemitism"\footnote{\url{https://decoding-antisemitism.eu/}} calls itself an ``AI-driven Study on Hate Speech and Imagery Online", and already produced five published reports on the subject. The project specifically aims at linking national or international events reported in the traditional media to antisemitic online social media discussions. 


\subsection{Alternatives to automated hate speech detection}
In \cite{Parker2023IsHS}, the authors question whether the way in which hate speech detection has been handled by the machine learning community is the way forward, or whether hate speech detection is a lot more complex than previously assumed by the researchers who labeled data sets and applied classifiers to them. Furthermore, 
the authors note that some hateful content may occur without the use of well-known slurs and that on top of it all, the nature of hate speech is constantly evolving. 

In contrast to previous studies, our work takes these observations into consideration and focuses on identifying emerging, potentially coded terms related to antisemitism using NLP methods. There has been a lack of rigorous research in finding emerging antisemitic coded terms that can lead to the detection of hate speech and, perhaps, subsequently, to the prevention of hate crimes. This paper aims to bridge this gap and provide an approach for the detection of emerging antisemitic, sometimes coded, terminology used on extremist social media platforms.

\section{Data Preparation} \label{tab:fw}
\begin{figure}
    \includegraphics[width=3in]{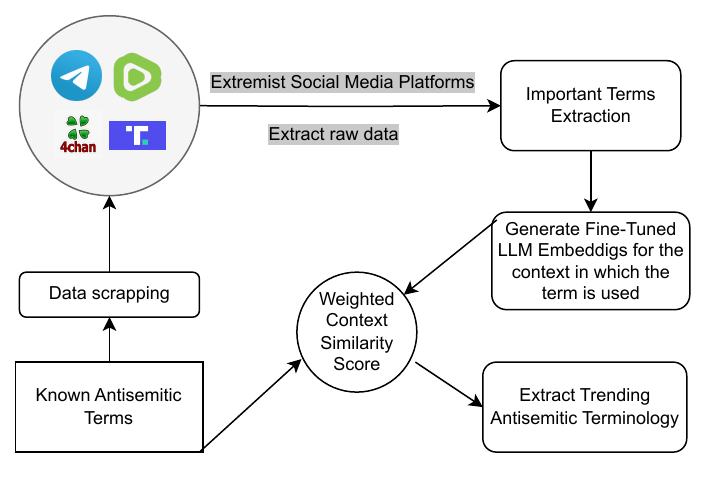}
    \caption{Emergent Coded Antisemitic Terminology Extraction Pipeline}    
    \label{tab:flow_diagram}
\end{figure}

This study is part of a large multi-disciplinary project 
sponsored by our institution which, simultaneously, collects and analyzes the use of coded language to express antisemitic sentiment in lightly moderated social media platforms typically preferred by individuals with extremist tendencies and studies the migration of this language from these extremist platforms to the general population. The overall project includes a data team, a population team, and a software team which collaborate closely and work in parallel. 
The pipeline illustrating our proposed methodology is shown in Figure~\ref{tab:flow_diagram}. 

\subsection{Dataset} \label{tab:methodology}
The project is constantly evolving, though for this study, we considered the first delivery of the data curated by the data team in June 2023. 
The data team's objectives concerning this study was to analyze the usage of antisemitic terms. 
We describe the data gathering and cleanup methodologies summarized by the 3 leftmost components in Figure~\ref{tab:flow_diagram}. 

\subsubsection{Data Scraping and Labeling} \label{tab:data_and_seed_words}

To build the corpus, the data team analyzed antisemitic social media posts from various extremist social media platforms including Discuss, Telegram, Minds, and GETTR. It used antisemitic 
expressions 
obtained from the previously mentioned American Jewish Committee (AJC) Translate Hate Glossary as well as the Southern Poverty Law Center (SPLC) 
to collect social media posts. This collection effort was facilitated by Pyrra\footnote{\url{https://www.pyrratech.com/}}, a private software company that allows its users to scrape posts from alt-social media platforms according to a list of seed terms. The data team considered the 46 seed expressions available from the AJC Glossary at the time as well as the term ``Cultural Marxism", discussed in a SPLC article\footnote{\url{https://www.splcenter.org/fighting-hate/intelligence-report/2003/cultural-marxism-catching}} and chose 16 of them to make the process tractable. It analyzed the 659 retrieved posts related to these seed expressions to determine whether the post was antisemitic or not.\footnote{A copy of the coding statement is available upon request.} The 16 terms used in the subset were selected based on their potential to reveal posts that had emerging new antisemitic terms in them. The list of seed words used is: \textit{Cabal, Cosmopolitan Elite, Cultural Marxism, Deicide, The Goyim Know, Holocough, Jewish Capitalist, Jewish Communist, Jew Down, Jewish Lobby, New World Order, Not the Real Jews, Rothschild, Soros, Zionist,} and \textit{Zionist Occupied Government}. 
%
Since the distribution of posts with respect to each seed expression is not ideal, though the software team used all the posts retrieved from the 16 seed expressions, 
it used only the seed expressions with at least 5 posts related to them to conduct its analysis. 
The terms dropped from the list according to this criterion are \textit{Jew Down} and \textit{Cosmopolitan elite}, leaving us with 14 seed words for the remainder of the study.

\subsubsection{Preprocessing}
Text preprocessing is a critical step in Natural Language Processing (NLP). It involves transforming raw text data into a format that can be easily analyzed by machine learning algorithms. The preprocessing steps usually used involve several techniques, such as tokenization, stop word removal, stemming, and lemmatization \cite{mustafa2017early}. 
During the first phase of cleaning the corpus, we removed the \texttt{urls} and lower-cased all the posts to normalize them. This initial procedure was followed by 
stop words removal. Then we lemmatized the text to get a single root form for each word prior to passing it on to the coded antisemitic terms extraction process, which will be discussed in the next section.
Bigrams and trigrams were formed by running two- and three- word windows through all the posts.\footnote{We also considered unigrams but were not able to filter them effectively using our current methodology. Their treatment was left for future work.}  It was important to filter out badly-formed expressions obtained through that approach. In particular, we decided to include bigrams and trigrams that only contain nouns, proper nouns, adjectives, and verbs, since others were judged less relevant to our quest.Since the emphasis of this study is on the novel proposed extraction process discussed next, we did not experiment with the various pre-processing techniques suggested in the literature on hate speech for social media posts \cite{Glazkova2023ACO}. It was left for future work.  
\section{Coded Antisemitic Terms Extraction Approach} \label{approach}
As previously mentioned, the purpose of this study is the extraction of \textit{emerging coded antisemitic terms}. In order to carry out this goal, we designed a method for operationalizing each term of that expression. That operationalization and the linking of its resulting components into a functional system constitute the main contribution of this work. The purpose of this section is to discuss the process. To begin with, we consider each word in the \textit{emerging coded antisemitic terms} expression and give it the specific meaning shown below.  
\begin{itemize}

    \item \textbf{Terms:} the extracted expressions are limited to \textit{grammatically consistent} bigrams and trigrams; they have to be \textit{relevant} enough to the documents in which they appear and appear \textit{frequently} enough in the corpus.
    \item \textbf{Antisemitic: } the candidate expressions have to be \textit{semantically related} to antisemitic discourse.
    \item \textbf{Coded: } antisemitic expressions that contain terms relating to obvious Jewish concepts are removed.
    \item \textbf{Emerging: } already known coded antisemitic expressions are removed in order to 
    concentrate on new terminology. 
\end{itemize}
These operations are divided into two phases. In Phase 1, we address the extraction of emerging coded terminology without worrying about its semantic relation to antisemitism. In Phase 2, we address semantics using large language models. Phase 1 is represented by the ``Important Terms Extraction" component in Figure~\ref{tab:flow_diagram}. Phase 2 is represented by the combination of the LLM Generation, Similarity Scoring, Antisemitic Terminology Extraction, and Monitoring components 
in Figure~\ref{tab:flow_diagram}. Both phases of the pipeline are implemented using two approaches: a standard solution and an advanced solution. We subsequently test all four combinations, yielding a baseline approach composed of two standard solutions, two hybrid approaches composed of one standard and one advanced solution, and one advanced approach composed of two advanced solutions.

\subsection{\underline{Phase 1: Emerging Coded Trending Terms Extraction}} \label{tab: phase_1}
For the first part of Phase 1, the extraction of trending terms, we explore the use of off-the-shelf NLP tools for our standard solution and then propose our advanced solution that combines tf-idf and frequency. 
Once the trending terms are extracted, we propose a strategy to remove non-emerging and non-coded terms from the list of extracted terms. This strategy is applied to both the standard and advanced solutions.
\subsubsection{\underline{Standard Solution:} Trending Terms Extraction using Concordance and Collocation tools}
In this first attempt at trending terms extraction, we use traditional NLP techniques to extract bi-grams and tri-grams using concordance and collocation algorithms from the NLTK Toolkit\cite{loper2002nltk}.  Concordance is a technique that provides a comprehensive view of how a given term appears in a corpus. Using this approach, we use the 14 seed terms from Section \ref{tab:data_and_seed_words} for analyzing patterns and gaining insights into language usage. For each occurrence of a seed term, this approach provides the surrounding words context. We use default settings for the extraction of context. Next, using collocation, we find the most frequent bi-grams and tri-grams in the collected contexts. Collocation is a technique that finds a meaningful combination of words from a corpus that are semantically coherent. Different statistical measures can be used to detect collocations including frequency, pointwise mutual information (PMI), and log-likelihood ratio (LLR) among others. We use frequency, here since that is the measure also used in the advanced approach 
In the future, we plan to experiment with other statistical measures for both approaches. The standard approach yielded 126 trending terms. 

\subsubsection{\underline{Advanced Solution:} Trending Terms Extraction using TF-IDF and Frequency}
 Our proposed advanced approach is presented in Algorithm 1 which uses TF-IDF feature-weighting \cite{ramos2003using} and frequency to extract trending terms. In a nutshell, this was done by selecting all the terms that obtained a TF-IDF value greater than a self-set threshold, listing these terms in decreasing order of frequency, and selecting the top 200 terms from the list.\footnote{We assume that at least 200 terms had a TF-IDF value larger than the self-set threshold.} When the same term appeared in several documents, the highest TF-IDF value it received was retained 
 Algorithms 1 shows the approach that was followed in detail. The algorithm is explained line by line next. 

\begin{algorithm}[h]
	\caption{Trending terms extraction}
        \label{alg:1}
	\begin{algorithmic}[1]
            \State Initialize Trending\_terms  \Comment{Stores top \textit{200} trending terms}
            \State Initialize $D_s$ \Comment{Stores terms' highest TF-IDF scores (s)}
             \State Initialize $D_f$ \Comment{Stores terms' values and frequencies}
             \State Set {\it T}\Comment{Stores all the vocabulary terms (value).} 
            \State Set {\it F} \Comment{Stores the frequency of each term in the corpus.} 
            \State Calculate the TF-IDF scores for each term in each document and store them in matrix $\mathbf{W}\in \mathbb{R}^{d\times v}$, where $d$ denotes the number of documents and $v$ denotes the vocabulary size.
            
            
		\For {each term $t$ in ${\bf T}$}
            \For{each row in $\mathbf{W}$}\Comment{Each row is a document }
        \State $D_s[t] \gets max(D_s[t], W[row, t]) $ \Comment{Finds t's highest TF-IDF  score, $s$, across all documents}
            \EndFor
		\EndFor
            \State $\delta \gets  Average(D_s)$ \Comment{$\delta$ is the average of all $s$'s}
        \State {i = 1}   
		\For  { each t in $D_s$}
		\If{$D_s[t]\geq \delta$} 
            \State $D_f[i] \gets  (T[t], F[t])$ \Comment{If t's highest TF-IDF score is larger than threshold $\delta$, store t's value and frequency in $D_f$}
            \State i = i + 1
            \EndIf
            \EndFor
            \State Sort $D_f$ in descending order of frequency
            \For{$i=1,2,\ldots, 200$ }
            \State $Trending\_terms \gets D_f[i][T]$ \Comment{Store the most frequent terms in Trending\_terms (drop the frequencies)}
            
            \EndFor
        \end{algorithmic}
\end{algorithm}


The algorithm begins by initializing the \texttt{Trending\_terms} list which is the list that will return the 200 bigrams and trigrams (terms) that received the highest combination of TF-IDF and Frequency scores.    
%
 Next, 
 $D_s$ and $D_f$ are also initialized. $D_s$ will be used to store all the terms' highest TF-IDF scores, whereas $D_f$ will store all the terms and their associated frequencies. Since terms are subsequently referred to according to their indices, $T$, which is set next, serves as the reference vector that associates an index with the actual value of the term (i.e., the actual bigram or trigram). Next, the frequency of each term in the corpus is calculated and saved in vector F. The TF-IDF values obtained for each unique term and each document are then calculated and placed in the matrix $\mathbf{W}$ of size \texttt{d x v}  where \texttt{d} represents the number of documents whereas, \texttt{v} is the number of terms. 
%

On lines 7-11, we take each term in matrix $\mathbf{W}$ and find the highest score across all the rows (documents) of the matrix and store it in \texttt{$D_s$}. 
To remove the less relevant terms, we compute the average of all the values in \texttt{$D_s$} and use this value, $\delta$, as a threshold. This allows us to consider only the terms with TF-IDF values greater than the average value of all the scores in \texttt{$D_s$}.  
$\delta$ is calculated on line 12. On lines 13-19, we check if the terms' TF-IDF value is greater than $\delta$. 
If so, we save the terms' values and their frequencies in \texttt{$D_f$}. 
Finally, on line 20, we sort the terms in \texttt{$D_f$} in descending order of their frequency values and select the top 200 terms, 
storing them in \texttt{Trending\_terms}. 
%
%
%

\subsubsection{\underline{Removal Strategy}: Redundant, Non-Emergent and Non-Coded terms Removal}
Once the list of 
most trending terms have been extracted using either the standard or advanced solution, three categories of terms are removed from it. 
First, 
as we consider expressions that are both bigrams and tri-grams, there is a possibility of encountering bigrams within trigrams. Such redundant bigrams are removed from the list of expressions. 
Next, we remove the terms that have occurred earlier. For now, this corresponds to the original list of 16 seed words used to retrieve the posts. In the future, this list will grow as we intend to use the system continuously, using newly discovered terms of interest as new seed terms. Lastly, the terms that are considered non-coded are removed. These correspond to 
terms that contain words that obviously pertain to Jewish themes. The list of words currently used includes jew, jewish, kike, and zionist. Expressions that include these words either as stand-alone words or embedded within other words are removed. 
After the removal phase is applied, we are left with 52 and 94 trending terms for the standard and advanced trending term extraction solutions, respectively.

\subsection{Phase 2: Embeddings and Comparisons}
Though the bigrams and trigrams extracted in the previous section are known to be trending, their semantics are unknown and, in particular, there is no information as to whether or not these terms are antisemitic. To find out which of these trending expressions are antisemitic, we 
compare the context in which they are used to the context in which the known antisemitic expressions are used. If a trending term appears in contexts similar to those in which seed expressions occur, it will be deemed antisemitic. Otherwise, it will be discarded as non-antisemitic. 
To compute embeddings for the trending and seed terms, we begin by fine-tuning BERT. Since BERT was not specifically trained on instances of hate speech or antisemitism, we fined-tuned it with additional data collected using the same seed expressions as before (since time elapsed between the original collection and the new collection, more posts were available for this exercise). This fine-tuned version of BERT is then used to generate contextual embeddings for both the trending terms discovered in the last section and the seed terms used to extract posts. We present the details of BERT's fine-tuning followed by the standard and advanced embedding solutions we implemented. 

\subsubsection{Fine-tuning the BERT model}
The generalized BERT model does not possess domain-specific vocabulary, thus it is not capable of handling coded hate speech such as antisemitism. Indeed, when such out-of-vocabulary terms occur, they get broken down into smaller tokens for which embeddings are generated. These are treated as rare tokens, yielding unsatisfactory results. 
To avoid this issue, we fine-tune the BERT model using an additional 56K posts extracted using the same seed words as before on Pyrra. We, thus, extend BERT's vocabulary from 30k to 55k tokens, and fine-tune it using the Masked Language Modeling (MLM) approach. MLM is a pre-training approach that masks a few tokens. The model is subsequently trained to predict the masked tokens from the words that surround them.\footnote{\url{https://huggingface.co/learn/nlp-course/chapter7/3}}

\subsubsection{Comparing Trending Terms to Seed Terms}
To differentiate between antisemitic and non-antisemitic terms during Phase 2, we compare the trending terms' embeddings to the 
seed terms' embeddings using Cosine Similarity.
%
We generate two types of embeddings following i) the standard pre-truncate embedding method 
and ii) the advanced post-truncate embedding method. 
\begin{figure}
    \centering
    \includegraphics[width=1\linewidth]{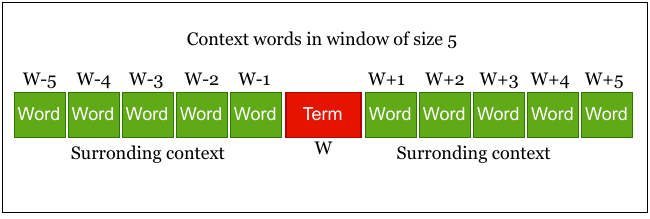}
    \caption{Pre-truncate embedding approach for a window of size 5.}
    \label{fig:context-window}
\end{figure}
In pre-truncate embedding, we truncate the post containing the term to be embedded \textit{prior to} embedding it. In post-truncate embedding, we embed the entire post containing the term of interest, and truncate the resulting embedding \textit{afterwards}.\footnote{Since we cannot embed posts exceeding 512 tokens, we turned large posts into multiple ones.}
\paragraph{\underline{Standard Solution:} Pre-truncate embeddings} \label{tab:contextofsurrondingwords}
In this approach, we consider context windows of 5 to 14 words, where the size of the windows  refers to the twin windows located before and after the term being embedded, respectively. We show an example of windows of size 5 in Figure \ref{fig:context-window}. 
Since the same term may be found in more than one post,  we concatenate all the embeddings extracted from fine-tuned BERT using the same window size and take their average. Embedding, here, refers to the \texttt{pooled layer} obtained from the 12 layers of the BERT architecture.
We follow the same procedure for all the trending terms we extracted and the 14 seed words retained in Section \ref{tab:data_and_seed_words}. 

Next, we determine the trending terms antisemitic nature using Algorithm~\ref{alg:2}. 
After some initialisations on lines 1-4, $S[tt]$, the ``similarity to antisemitism" value for trending term $tt$, is computed as follows: 
$tt$'s embedding is compared to each of the 14 seed terms (the $st$'s)'s embeddings using Cosine Similarity as described on lines 6-8. On line 9, the 14 resulting measurements are averaged and assigned to $S[tt]$ 
The process is repeated for each trending term (lines 5-10) and the median of all the $S[tt]$'s, $\gamma$, is calculated on line 11.  
 $\gamma$ is then used as our threshold for potential antisemitism on lines 12-18: if $S[tt]$ for trending term $tt$ is greater than $\gamma$, $tt$ will be given the partial label ``potentially antisemitic" ($TT\_PL\_w[tt]$ = 1).  Otherwise, it will be given the partial label ``probably not antisemitic" ($TT\_PL\_w[tt]$ = 0). (We used the median as it offered more flexibility than the mean.) 
 Algorithm~\ref{alg:2} is repeated 10 times, once for each window size $w$ considered. This yields 10 partial labels $TT\_PL\_w[tt]$, $w = 1 \dots 10$ for each term $tt$, and the final labeling for $tt$ is ``antisemitic" if $m$ out of the 10 partial labels are ``potentially antisemitic". It is ``not antisemitic", otherwise. The optimal value of $m$ was 7 for the pre-truncate case.
%
%

\begin{algorithm}[h]
	\caption{Comparing semantic similarity--window size $w$}
 \label{alg:2}
	\begin{algorithmic}[1]
            \State $Embeddings\_tt \gets \{et\_1,et\_2\dots, et\_n\}$\Comment{n pre- or post- truncate trending terms embeddings at window size $w$ }
            \State $Embeddings\_st \gets \{es\_1,es\_2\dots,es\_14\}$\Comment{14  pre- or post- truncate seed words embeddings at window size $w$}
            \State Initialize TT\_PL\_w. TT\_PL\_w will store the n trending terms \&  predicted antisemitic label for window size $w$.
            \State Initialize $S$. $S$ will store the average semantic score for each trending term at window size $w$.
		\For {each \textit{tt} in ${\bf Embeddings\_tt}$}
            \For{each \textit{st} in ${\bf Embeddings\_st}$}
            \State $tt\_scores[tt] \gets Sim(et\_tt, es\_st)$ \Comment{Cosine Sim}
            \EndFor
            \State {$S[tt] \gets Average({tt\_scores[tt]}$)} \Comment{Average all the 14 semantic scores between tt and all the st's} 
    \EndFor
            \State {$\gamma \gets Median(S)$} \Comment{$\gamma$ is the median of all the scores}
        \For {each \textit{tt} in ${\bf S}$}
            \If{$S[tt] > \gamma$}\Comment{check if score greater than $\gamma$}
            \State {$TT\_PL\_w[tt] \gets 1 $}\Comment{if score greater than $\gamma$}
            \Else
            \State {$TT\_PL\_w[tt] \gets 0 $}\Comment{if score less than $\gamma$}
            \EndIf
        \EndFor
        \end{algorithmic}
\end{algorithm}
\begin{figure}
    \centering
    \includegraphics[width=1\linewidth]{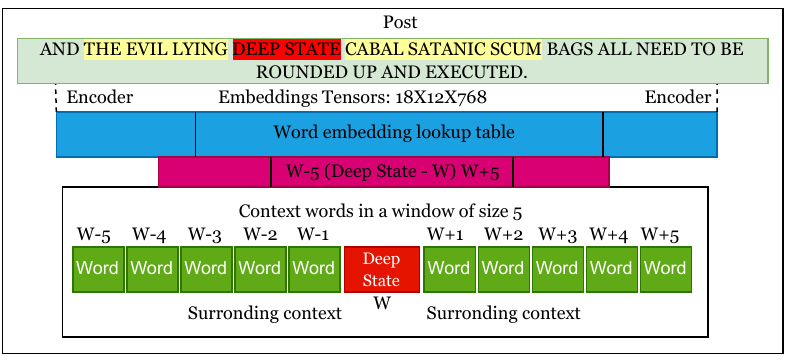}
    \caption{Post-truncate embedding approach for a window of size 5.}
    \label{fig:complete-context-window}
\end{figure}

\paragraph{\underline{Advanced Solution:} Post-truncate embeddings}
In this approach, we begin by embedding each complete post using fine-tuned BERT. 
The approach is illustrated in Figure \ref{fig:complete-context-window} for the 18-word post \texttt{AND THE EVIL LYING \textbf{DEEP STATE} CABAL SATANIC SCUM BAGS ALL NEED TO BE ROUNDED UP AND EXECUTED.} 
This yields an $18$ x $12$ x $768$ tensor representing the total number of words in the post, the total number of encoding layers, and their dimension. 
This embedding can be thought of as a word embeddings lookup table that provides complete context for each post.\footnote{We assume that each word in the post has a token id in Bert's vocabulary.}
Once this embedding is constructed, we follow the same procedure described in 
Section \ref{tab:contextofsurrondingwords} except for the fact that 
we now extract word-level contextual embeddings from the lookup table (see Figure \ref{fig:complete-context-window}). The advantage of this approach over the previous one is that it builds more informed embeddings given its use of a complete rather than partial context. Please note that there are three additional differences between the standard and the advanced approach: in the standard approach, we used context window sizes between 5 and 14 while in the advanced approach, we used context window sizes between 1 and 10. That is because a window of 1 word does not convey much information in the standard approach whereas it does in the advanced approach. As a result, we started at size 5 in the standard approach and 1 in the advanced approach and used 10 different window sizes in each case. Furthermore, in the advanced approach, the embeddings are generated by averaging the final encoder layer of BERT rather than using the pooled layer since that yielded better results. Finally, the optimal value for $m$ in the advanced approach was 9 rather than 7. 

%


\section{Results and Discussion} \label{tab:results_discussion}
The purpose of our study was to design a methodology for extracting emerging coded antisemitic terminology from online posts appearing on social media platforms often used by extremist groups. We proposed a pipeline to implement this methodology and instantiated this pipeline with standard and advanced components. 
The difficult part of our evaluation is the assessment of whether the approach yields a significant number of terms and whether these terms can, indeed 
 in some contexts, have an antisemitic connotation. In order to answer these questions, we created a gold standard and tested our results according to it. 
%
%
\subsection{Construction of a gold standard:} 
The gold standard we created uses two complementary methodologies. One for the  terms already familiar to the community that fights antisemitism, and the other, for the terms unknown or not yet catalogued by that community.\footnote{In this paper, we created a prototype system based on the seed words provided to us by the data team. These seed words are only a small subset of the already known coded antisemitic terms. As a result, some of the emergent terms discovered by our system are emergent vis-a-vis the system's knowledge but not vis-a-vis the broader current knowledge. Discovering terms known to the community but not known by the system constitutes a useful proof of concept. The discovery of terms not currently known by the community constitutes an added demonstration of the worth of the approach.}  

{\it \underline{Known Terms}} For the first category, we simply compiled a general glossary from  three existing sources: the Institute for Curriculum Services' Glossary spanning the history of European Antisemitism, which we took in its entirety;  the American Jewish Committee ``Translate Hate" glossary which we also used in its entirety (prior to its recent expansion from 46 to 70 terms) and portions of the Glossary of Terms and Acronyms constructed by the R2Pris project on Radicalization and violent extremism. Since this last source encompassed hatred of different types, for this specific study, we restricted ourselves to the terms whose composition or definition included a known antisemitic term (e.g., nazi, Aryan, anti-semitic, Ku Klux Klan, SS, Swastika, Fascism, White Supremacist, 
etc.).\footnote{ The sources we used can be found at the following websites: \url{https://bit.ly/45kEtYB}; \url{https://bit.ly/3MIjKpt}; and \url{http://www.r2pris.org/glossary.html} }
{\\\it \underline{New Terms}} The new terms are the terms that do not appear in the glossaries just mentioned 
and that need to be manually verified through an internet search. We used the following systematic procedure to assign ground labels to new terms:


\begin{itemize}
    \item Each extracted term not found in the glossary compilation was searched for on Google in two ways: the term alone or together with ``+ antisemitism" added to the search.
    \item The documents retrieved on the first page of the Google browser for both searches were examined for references to antisemitism.
    \item If, based on this analysis, the term was found to be associated with antisemitism (e.g., ``deep state" was found to be associated with a conspiracy theory against the Jews), it was coded as antisemitic in our gold standard database. If, on the other hand, the term did not carry any clear meaning (e.g., ``late 20th") or was not associated with antisemitism (e.g., ``new york city"), it was coded as not antisemitic in our gold standard database.
\end{itemize}
{\it \underline{Qualitative evaluation}}
We conducted two types of qualitative evaluation. The first one simply consisted of observing the terms extracted by the approach to assess whether they made sense when taken out of context. The second one can be thought of as a sanity check. For terms extracted and labeled as either antisemitic or not, 
we went back to the 
the posts from which the term was extracted to assess whether, within the context of the post, it was used in an antisemitic way or not. Though we do not use these qualitative assessments in our quantitative evaluation, 
we show examples of the different situations that arose in terms of agreement or disagreement between our system and our gold standard. 

\subsection{Results}
\begin{table*}
    \centering
    \caption{Accuracy, Precision, Recall and F-score using the four versions of our pipeline. }
    \begin{tabular}{|lccccc|}
         \hline
\textbf{Model+Embedding} & \textbf{Approach Type} &\textbf{Accuracy} & \textbf{Precision}  &  \textbf{Recall} & \textbf{F-score} \\
\hline\hline
colloc-pretrunc & standard & 0.74      & 0.34  & 1  & 0.51 \\ \hline
colloc-posttrunc & hybrid &0.76       & 0.36  & 1  & 0.53 \\ \hline 
tfidf-pretrunc & hybrid &0.67       & 0.47  & 0.55  & 0.51 \\ \hline
tfidf-posttrunc & advanced &\textbf{0.80}       & \textbf{0.63}  & \textbf{0.83}  & \textbf{0.72} \\ \hline 



    \end{tabular}
    
    \label{tab:results}
\end{table*}

{\it \underline{Quantitative Results:}} 
We tested four different versions of our proposed pipeline, by combining the standard and advanced solutions proposed for trending term extraction with the standard and advanced solutions proposed for term embedding. These combinations resulted in one standard, two hybrid, and one advanced implementation. Table~\ref{tab:results} lists the results obtained by concordance + collocation followed by pre-truncation embedding (colloc-pretrunc) or post-truncation embedding (colloc-posttrunc); and those obtained by tfidf + frequency followed by pre-truncation embedding (tfidf-pretrunc) or post-truncation embedding (tfidf-posttrunc). The results were obtained using our gold standard labels. 
The approach using the two advanced components stands out as the absolute winner: tfidf-posttrunc, although the results for all four methods, including tfidf-posttrunc, show a higher level of recall than precision. Future work will attempt to improve all these metrics scores, with a focus on precision so as not to unduly label terms as antisemitic when they are, in fact, benign. When comparing the numbers in Table~\ref{tab:results}, it is important to note that the number and type of terms retrieved differ between the two term extraction processes, colloc and tfidf. While colloc extracted 52 terms of which only 7 were truly antisemitic, tfidf extracted 94 of which 29 were truly antisemitic. The recall of 1 obtained by the two colloc-based methods, thus means that both pretrunc and posttrunc were able to identify these 7 antisemitic terms. Their low level of precision, however, suggests that they are too liberal in their labeling of terms as antisemitic.  

\begin{table*}
    \centering
    \caption{List of trending terms that are predicted antisemitic by the most advanced version of the pipeline.}
    \begin{tabular}{|llllllll|}
         \hline
         \textbf{{\color{red} \underline{False Positives}}} && \textbf{\underline{Known Terms}} && \textbf{\color{blue} \underline{New Terms}}  && \textbf{\color{purple}\underline{Neutral}}& \\\hline
 {\color{red}{plain sight}}&{\color{red}{german people}} & {{white genocide}} & {{interest groups}} &  {\color{blue}{FEMA camps}}&  {\color{blue}{color revolution}} & {\color{purple}{end game}}& {\color{purple}{world war}} \\
\hline
 {\color{red}{new york city}}& {\color{red}{big part}}  & {nostra aetate}& {federal reserve}  & {\color{blue}{central bank}} & {\color{blue}{critical race theory}}  & {\color{purple}{western civilization}}& {\color{purple}{democrat party}} \\ \hline 
 

    \end{tabular}
    
    \label{tab:trending_terms_results}
\end{table*}

{\it \underline{Qualitative Results:}} Our qualitative evaluation was applied to the version of our pipeline that obtained the best results: tfidf-posttrunc, i.e., the advanced version. Table~\ref{tab:trending_terms_results} shows some of the terms extracted by that version. The terms in red correspond to terms incorrectly classified as antisemitic with no good explanation; those in black are correctly classified as antisemitic as they correspond to our {\it {Known Terms}}; those in blue were verified to be antisemitic as they correspond to our {\it {New Terms}}; and those in purple were incorrectly classified as antisemitic, although the context in which they arise is clearly antisemitic. As discussed below, we call these terms {\it {Neutral} (in an antisemitic context)}.

{\it \underline{Sanity Check:}} 
In Table \ref{tab:tweets}, we show a few sample posts containing the following trending antisemitic terms discovered by tfifd-posttrunc: \texttt{Interest groups}, \texttt{White Genocide}, \texttt{Deep state}. 
Each of these terms had an entry in the antisemitic glossary compilation described earlier. 
For instance, 
\texttt{White Genocide}, refers to a conspiracy theory rooted in white supremacist ideology, claiming that there is an intentional effort by Jews to destroy the white race through immigration, mixed-racial marriage,  LGBTQ+ identification, etc. 
\begin{table*}
\setlength{\tabcolsep}{2pt} 
\caption{Posts on social media with automatically labeled 
antisemitic coded terms 
as per the most advanced version of the pipeline.} 
\label{tab:tweets}
\centering
\begin{tabular}{|l|l|l| p{12cm}|}
\hline
Coded Term      & Status   & Website & Post \\ \hline
Interest groups  & Known Term & Minds & the united states government is controlled by {\bf interest groups} that are only seeking to enlarge their own power. the us government does not represent the will of the citizenry, and condemning it is not a condemnation on the principles of freedom, democracy, etc.the usa is being set up to fail.the rootless cosmopolitan elite have been constructing elaborate safehouses for decades in preparation for this.    \\ \hline
Deep State & Known Term & 4chan & US { \bf deep state} MIGApede detected. The real {\bf deep state} is the Jewish lobby.    \\ \hline
White Genocide & Known Term & Truth Social & Rotten Eggs - Dr. Reiner Fuellmich and Whitney Webb! Vatican Pro-Abortion- False Prophet Francis Owned By New World Order! Jacob's Trouble = { \bf White Genocide}! Pandemic Of The Double Dosed.Inflation Spiking, More Lockdowns, The Worst Is Yet To Come! \\ \hline
{\color{purple} End Game}/{\color{blue}FEMA camps} & {\color{purple}Neutral}/{\color{blue}New Term} & Truth Social &FEMA  is not a good thing! {\bf \color{blue} FEMA camps} are concentration camps. {\bf \color{blue} FEMA camps} are the {\bf \color{purple} end game} of the New World Order\\\hline
{\color{red} Big Part} & {\color{red}False Positive}& 4chan &all turds need to be deported from the West. turds are brown MENA sunni muslim garbage. they are a {\bf \color{red}big part} of the non-white invasion. many of the turkish Iraqi and syrian immigrants who rape women and girls are actually ethnic turds. turds are also zionists and turdistan is a base for israeli ops. imagine sympathizing with these zio-muslim invaders.\\\hline
\end{tabular}
\end{table*}
Table \ref{tab:tweets} also shows an instance of a new term ---\texttt{FEMA camps}. This corresponds to a conspiracy theory where FEMA is believed to plan the incarceration and possible execution of US citizens in favor of the establishment of a New World Order, one of our seed words which often refers to the establishment of a new form of government controlled by a Jewish elite.
On the other hand, during the process of extracting coded antisemitic terms, some terms were labeled as antisemitic despite the fact that they do not appear in our gold standard. In certain cases, that represents an outright mistake like in the case of \texttt{Big Part} in Table~\ref{tab:tweets} where the context is certainly racist, but not specifically antisemitic, though antisemitism is part of the post, but in other situations, a case could be made for the antisemitic label. For example, our approach predicts \texttt{End game} as a coded antisemitic term, even though we did not find any reason for it in our glossary or internet search. A look at the post in which the term appeared (Table~\ref{tab:tweets}) helps us understand how the antisemitic context of the post that includes the terms ``concentration camps" and "new world order" led the system to mislabel it.  
We conclude that, in such cases, our approach is extracting the right term according to the context, but the term should be considered {\textit{Neutral} \textit{(in an antisemitic context)}} rather than antisemitic. 
\subsection{Discussion}
Though we assume that our approach could still be refined, we note that the results obtained by the most successful version of our system are encouraging, suggesting the viability of our hypothesis that emergent coded terms could be discovered automatically using distance measures in embedding spaces. The sanity checks suggest that the terms identified by our approach are, usually, warranted as the context shown in the posts attests to the antisemitic nature of the way in which the identified terms are used. These checks also point to the errors made by the system and will help us improve our results. 
We also believe that our approach could have important practical uses. After being vetted by a human team, the emergent coded terminology it discovers could be input to the moderating algorithms used by social media platforms to discover problematic discourse or users currently avoiding discovery. 
\section{Conclusion} \label{tab:conclusion}
This paper proposes an approach for detecting the emergence of new antisemitic coded terminology 
%
%
which offers a valuable resource in combating online antisemitism and contributes to the ongoing efforts to create safer and more inclusive online spaces. We achieve an accuracy of 80\% and F-Score of 72\% in extracting antisemitic terms using this approach which relies on NLP techniques including POS tagging, TF-IDF, and 
Fined-tuned large language models such as BERT.
%
%
In the future, we intend to refine our semantic similarity technique by exploring other deep learning and large language model approaches and their various parameter combinations. 
Similarly, we will experiment with different types of text pre-processing approaches to deal specifically with hate-speech and social media text. This will be done in the context of a lifelong-learning setting where the trending terms discovered will be used as input to the data scraping component in the following iteration. We also intend to create a more user-friendly version that will be convenient for people working in this space. Finally, our goal is to extend this study to hateful terminology against other minority groups.

\bibliographystyle{ieeetr}
\textbf{\bibliography{bibliography}
}

\end{document}